\documentclass[runningheads]{llncs}
\usepackage{subfigure}
\usepackage{graphicx}
\usepackage{gensymb}
\usepackage{tabularx,booktabs}
\usepackage{multirow}
\newcolumntype{Y}{>{\centering\arraybackslash}X}
\begin{document}
\title{Vox2Vox: 3D-GAN for Brain Tumour Segmentation}
%
\author{Marco Domenico Cirillo\inst{1,2} \and
David Abramian\inst{1,2} \and
Anders Eklund\inst{1,2,3}}
\authorrunning{M.D. Cirillo et al.}

%
\institute{Department of Biomedical Engineering, Link\"oping University, Link\"oping, Sweden \and
Center for Medical Image Science and Visualization, Link\"oping University, Link\"oping, Sweden \and
Division of Statistics and Machine learning, Department of Computer and Information Science, Link\"oping University, Link\"oping, Sweden
\email{\{marco.domenico.cirillo, david.abramian, anders.eklund\}@liu.se}}

\maketitle              
\begin{abstract}
Gliomas are the most common primary brain malignancies, with different degrees of aggressiveness, variable prognosis and various heterogeneous histological sub-regions, i.e., peritumoral edema, necrotic core, enhancing and non-enhancing tumour core. Although brain tumours can easily be detected using multi-modal MRI, accurate tumor segmentation is a challenging task. Hence, using the data provided by the BraTS Challenge 2020, we propose a 3D volume-to-volume Generative Adversarial Network for segmentation of brain tumours. The model, called Vox2Vox, generates realistic segmentation outputs from multi-channel 3D MR images, segmenting the whole, core and enhancing tumor with mean values of 87.20\%, 81.14\%, and 78.67\% as dice scores and 6.44mm, 24.36mm, and 18.95mm for Hausdorff distance 95 percentile for the BraTS testing set after ensembling 10 Vox2Vox models obtained with a 10-fold cross-validation.

\end{abstract}

\section{Introduction}
Gliomas are the most frequent intrinsic tumours of the central nervous system. Based on the
presence or absence of marked mitotic activity, necrosis and ﬂorid microvascular proliferation, a malignancy grade, WHO grade II, III or IV \cite{wesseling}, is assigned. Gliomas with WHO grade II are also called low grade gliomas (LGG), whereas gliomas with higher WHO grade are called 
high grade gliomas (HGG). Although both these brain tumour types can easily be detected, they have a diffuse, infiltrative way of growing in the brain, and they exhibit peritumoural edema, such as an increase in water content in the area surrounding the tumour. This makes it arduous to define the tumour border by visual assessment, both in analysis and also during surgery \cite{blystad}.

For this reason, researchers recently started resorting to powerful techniques, able to segment complex objects and, in this way, guide the surgeons during the operation with a suitable accuracy. Indeed, machine learning \cite{polly} and especially deep learning \cite{isensee,kamnitsas,kong,mckinley,myronenko,topol} can provide state-of-the-art segmentation results.

\subsection{Related Works}
Nowadays generative adversarial networks (GANs) \cite{goodfellow} are gaining popularity in computer vision, since they can learn to synthesise virtually any type of image. Specifically, GANs can be used for style transfer \cite{gatys}, image synthesis from noise \cite{karras}, image to image translation \cite{isola}, and also image segmentation \cite{sato}. GANs have become especially popular in medical imaging \cite{yi} since medical imaging datasets are much smaller compared to general computer vision datasets such as ImageNet. Additionally, in medical imaging it is common to collect several image modalities for each subject before proceeding with the analysis, and, when this is not possible, CycleGAN introduced in \cite{zhu} can be used to synthesize the missing modalities.

GANs have also been used for medical image segmentation. Indeed,  Han \textit{et al.} in \cite{han} proposed a GAN to segment multiple spinal structures in MRIs; Li \textit{et al.} in \cite{li} developed a novel transfer-learning framework using a GAN for robust segmentation of different human epithelial type 2 (HEp-2) cells; Dong \textit{et al.} in \cite{dong} implemented a U-Net style GAN for accurate and timely organs-at-risk (OARs) segmentation; Nema \textit{et al.} in \cite{nema} designed a 2D GAN, called RescueNet, to segment brain tumours from MR images; etc. Anyhow, Yi \textit{et al.} \cite{yi} provide a complete and recent review of GANs applied in medicine.\\

Hence, inspired by these works and especially by the Pix2Pix GAN \cite{isola}, which can generate an image of type A from a paired image of type B, the aim of this project is to do 3D image segmentation using 3D Pix2Pix GAN, named Vox2Vox, to segment brain gliomas. While a normal convolutional neural network, such as U-Net \cite{ronneberger}, performs the segmentation pixel by pixel, or voxel by voxel, through maximizing a segmentation metric or metrics (i.e. dice score, intersection over union, etc), a GAN will also punish segmentation results that do not look realistic. Our hypothesis is that this can result in better segmentations.

\section{Method}
\subsection{Data}
\begin{figure}
    \centering
    \includegraphics[width=1\linewidth]{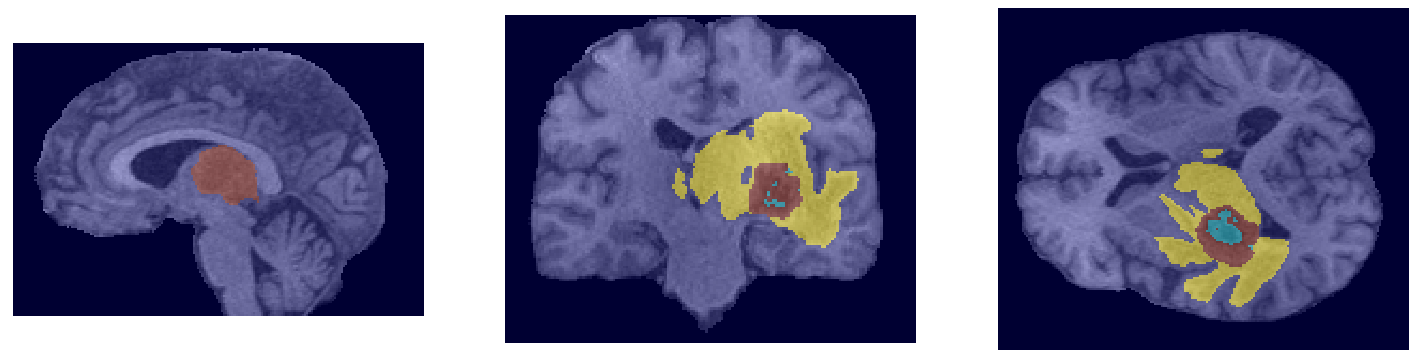}
    \caption{From left to right there is a T1 MR image in the sagittal, coronal and transverse plane overlapped with its true segmentation. Peritumoural edema (ED), necrotic and non-enhancing tumour core (NCR/NET), and GD-enhancing tumour (ET) are highlighted in yellow, red and cyan respectively.}
    \label{fig:data}
\end{figure}

The MR images used for this project are the Multimodal Brain tumour Segmentation Challenge (BraTS) 2020 training ones \cite{bakas3,bakas4,bakas1,bakas2,menze}. The BraTS 2020 training dataset contains MR volumes of shape \(240\times240\times155\) from 369 patients, and for each patient four types of MR images were collected: native (T1), post-contrast T1-weighted (T1Gd), T2-weighted (T2), and T2 Fluid Attenuated Inversion Recovery (FLAIR). The BraTS 2020 validation dataset contains MR volumes from 125 patients. The images were acquired from 19 different institutions with different clinical protocols. The training set was segmented manually, by one to four raters, following the same annotation protocol, and their annotations were approved by experienced neuro-radiologists, whereas no segmentation was provided for the validation set. Moreover, all data were co-registered to the same anatomical template, interpolated to the same resolution (1 mm\(^3\)) and skull-stripped. 
Figure \ref{fig:data} shows an example of one training T1 MR image overlapped with its true segmentation in the theree different planes.
\subsection{Image Pre-processing and Augmentation}
For each MR image intensity normalization is done per channel, whereas the background voxels are fixed to 0.  On the other hand, the grey-scale ground-truths are transformed into categorical, so each target has four channels, as the number of the classes to segment: background, peritumoural edema (ED), necrotic and non-enhancing tumour core (NCR/NET), and GD-enhancing tumour (ET) labeled with 0, 1, 2, 4 respectively. BraTS, moreover, released 125 and 166 additional volumes as validation and test sets, respectively.

Since these volumes are memory demanding, patch augmentation is applied to extract one sub-volume of \(128\times 128\times 128\) from each original volume. In this way, only 23.5\% of the whole training set is used in every training epoch. Moreover, in order to prevent the networks from overfitting and memorizing the exact details of the training images, random 3D flipping, random 3D rotations between 0\degree and 30\degree, power-law gamma intensity transformation (gain and gamma randomly chosen between [0.8-1.2]), elastic deformation with square deformation grid with displacements sampled from from a normal distribution with standard deviation 5 voxels \cite{ronneberger}, or a combination of these with probability 0.5 are applied as image augmentation techniques.

\subsection{Model Architecture}
\begin{figure}[h!]
  \centering
  \includegraphics[width=1\linewidth]{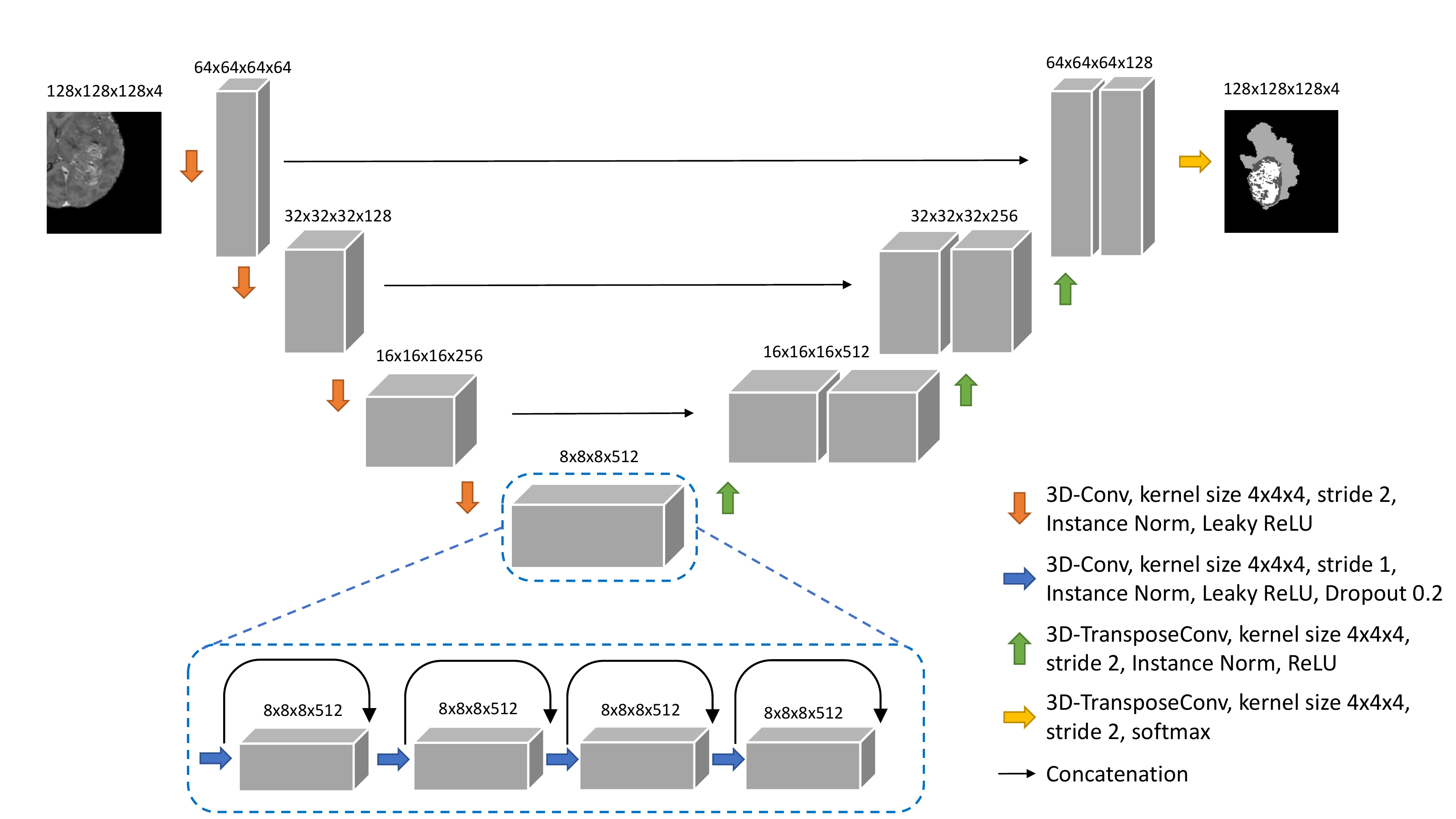}
  \caption{The generator model.}
  \label{fig:gen}
\end{figure}

\begin{figure}[h!]
  \centering
  \includegraphics[width=01\linewidth]{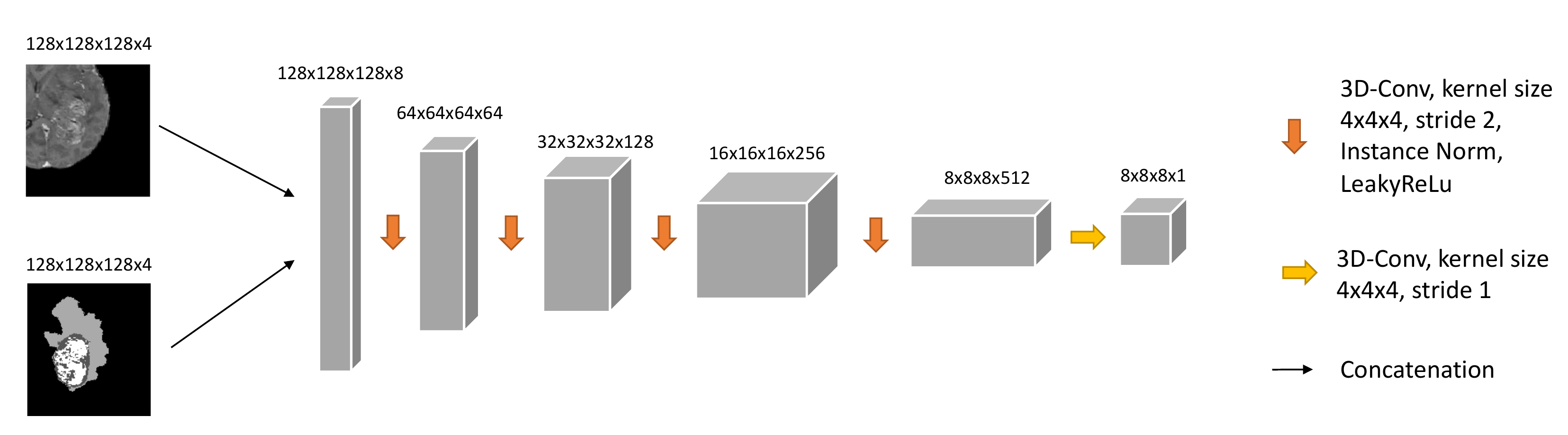}
  \caption{The discriminator model.}
  \label{fig:disc}
\end{figure}

The Vox2Vox model, as the Pix2Pix one \cite{isola}, consists of a generator and a discriminator. The generator, illustrated by Figure \ref{fig:gen}, is built with U-Net and Res-Net \cite{he2016deep} architecture style, see Figure \ref{fig:gen}; whereas the discriminator is build with PatchGAN \cite{isola} architecture style, see Figure \ref{fig:disc}. Following are the details of both model's architecture\footnote{In the model descriptions we use \textit{I}, \textit{E}, \textit{B}, \textit{D}, and \textit{O} to refer to Input(s), Encoder, Bottleneck, Decoder, and Output respectively.}. The generator consists of:
\begin{itemize}
    \item[\textit{I:}] a 3D image with 4 channels: T1, T2, T1Gd, and T2 FLAIR;
    \item[\textit{E:}] four down-sampling blocks, each of them made by 3D convolutions using kernel size 4, stride 2 and same padding, followed by instance normalization \cite{ulyanov} and Leaky ReLU activation function. The number of filters used at the first 3D convolution is 64 and at each down-sampling the number is doubled;
    \item[\textit{B:}] four residual blocks, each made by 3D convolutions using kernel size 4, stride 1 and same padding, followed by instance normalization and Leaky ReLU activation function. Every convolution-normalization-activation output is concatenated with the previous one;
    \item[\textit{D:}] three up-sampling blocks, each of them made by 3D transpose convolutions using kernel size 4 and stride 2, followed by instance normalization and ReLU activation. Each 3D convolution input is concatenated with the respective encoder output layer;
    \item[\textit{O:}] a 3D transpose convolution using 4 filters (as the number of the classes to segment), kernel size 4 and stride 2, followed by softmax activation function. The output has shape \(128\times128\times128\times4\) and constitutes the segmentation prediction for each class.
\end{itemize}
On the other hand, the discriminator consists of:
\begin{itemize}
    \item[\textit{I:}] the 3D image with 4 channels and its segmentation ground-truth or the generator's segmentation prediction;
    \item[\textit{E:}] same as the generator;
    \item[\textit{O:}] one 3D convolution using 1 filter, kernel size 4, stride 1 and same padding. The output has shape \(8\times8\times8\times1\) and constitutes the quality of the segmentation prediction created by the generator.
\end{itemize}

All 3D convolution and 3D transpose convolution layers use kernel size 4 (as in \cite{isola}), the He \textit{et al.} weight initialization method\cite{he}, and same padding. Moreover, all Leaky ReLU layers have slope coefficient 0.3.\\
\\
In order to reduce GPU memory consumption, we decided to used a single convolution layer in the encoder and decoder blocks, while a standard U-Net uses two convolutions layers per block. As our convolutions use stride 2, it means that the encoder/decoder will directly downsample/upsample the input volume a factor 2. In future work we will investigate if using two convolutions layers per blocks improves segmentation performance.

\subsection{Losses}
Since Vox2Vox contains two models, the generator and the discriminator, two loss functions are used. The discriminator loss, \(L_D\), is the sum of the \(L_2\) error of the discriminator output, \(D(\cdot, \cdot)\), between the original image \(x\) and the respective ground-truth \(y\) with a tensor of ones, and the \(L_2\) error of the discriminator output between the original image and the respective segmentation prediction \(\hat{y}\) given by the generator with a tensor of zeros, i.e.:
\begin{equation}
    L_{D} = L_2\left[D(x, y), \mathbf{1}\right] + L_2\left[D(x, \hat{y}), \mathbf{0}\right]\,,
\end{equation}

whereas, the generator loss, \(L_G\), is the sum of the \(L_2\) error of the discriminator output between the original image and the respective segmentation prediction given by the generator with a tensor of ones, and the generalized dice loss \cite{milletari,sudre}, \(GDL(\cdot, \cdot)\), between the ground-truth and the generator's output multiplied by the scalar weight coefficient \(\alpha\geq 0\), i.e.:
\begin{equation}
    L_{G} = \, L_2\left[D(x, \hat{y}), \mathbf{1}\right] + \alpha\, GDL\left(y, \hat{y}\right)\, .
    \label{eq:loss}
\end{equation}

By looking at Equation \ref{eq:loss}, it is easy to conclude that if \(\alpha=0\): Vox2Vox is a pure GAN and it minimizes only the unsupervised loss given by the discriminator; whereas if  \(\alpha\rightarrow \infty\): Vox2Vox ignores the discriminator, and behaves as a 3D U-Net with the architecture shown in Figure \ref{fig:gen}.

\subsection{Optimization and Regularization}
Both the generator and the discriminator are trained using the Adam optimizer \cite{kingma} with the parameters: \(\lambda = 2\cdot 10^{-4}\), \(\beta_1 = 0.5\), and \(\beta_2 = 0.999\). Dropout regularization with a dropout probability of 0.2 is used after each 3D convolutional operation in the generator's bottleneck (see Figure \ref{fig:gen}).
Moreover, as Yi \textit{et al.} reported in \cite{yi}, the discriminator loss helps the generator to guarantee the spatial consistency in the final results, behaving as a shape regularizer. In other words, the discriminator takes care that the generated brain segmentation looks realistic (i.e. like manual segmentations). In the end, we expect that Vox2Vox performs better with a trade-off \(\alpha\) which does not disregard completely the discriminator loss and, at the same time, does not disregard the generator either.

\subsection{Model Ensembling and Post-processing}
Model ensembling is a technique that combines the outputs of several models in order to obtain more robust predictions \cite{kamnitsas}. 
Hence, once the Vox2Vox model is built, \(M\)-fold cross-validation over the training data can be done, which results in \(M\) models.
Instead of performing the ensembling independently for each voxel, we propose a neighborhood averaging ensembling, by training a CNN that combines the outputs of the \(M\) Vox2Vox models. This CNN, called Ensembler, is simply made by just one 3D convolution, which takes as input a tensor with shape \((128\times128\times128\times4M)\), and returns a tensor with shape \((128\times128\times128\times4)\), using stride 1, kernel size 3 and softmax activation function. 
Since the input of our Ensembler are softmax outputs from each Vox2Vox model, the probabilities were zero-centered by subtracting 0.5 prior to training. In the end, the Ensembler is trained over 100 epochs, minimizing the generalized dice loss, and early stopping with a patience of 10 epoch (for the validation loss) is used.

Furthermore, since the \(M\) models are trained to detect 4 classes, it is normal that they sometimes detect a class which is not present for a subject and the ensembling follows such mis-segmentation. Therefore, post-processing can be useful in order to reduce mis-segmentation. Since the mis-segmentation normally results in small false positives, a cluster size threshold can be used to remove clusters smaller than a specific volume \(V\).





\section{Results}
The Vox2Vox model is implemented using Python 3.7, Tensorflow 2.1 and its Keras library. The model is trained and validated on sub-volumes of size \(128\times128\times128\) from 369 and 125 subjects respectively, using batch size 4, over 200 epochs on a computer equipped with 128 GB RAM and an Nvidia GeForce RTX 2080 Ti graphics card with 11 GB of memory.
Once the training is completed, the 166 test volumes are cropped in order to have shape \(160\times192\times128\). In this way, the testing set can be given as input to the fully convolutional Vox2Vox, because each axis is now divisible by \(2^4 = 16\), where 4 is the generator's and discriminator's depth.

Table \ref{tab:alphas} reports the dice and the Hausdorff distance 95 percentile scores for all the classes of interest varying the \(\alpha\) parameter in Equation \ref{eq:loss} for the training dataset. Note that the classes are reset as: whole tumour (WT = ET \(\cup\) ED \(\cup\) NCR/NET), tumour core (TC = ET \(\cup\) NCR/NET) and enhancing tumour (ET). It also clearly shows that the best trade-off for the \(\alpha\) parameter is 5, but at the same time that it cannot detect the enhancing tumor (ET) class properly over the whole training dataset. Anyway, it also shows that the discriminator helps to achieve good results, because the metrics obtained with \(\alpha=5\) are better than when the model only considers the generator loss (high values of \(\alpha\)) and also when the model is a pure GAN (\(\alpha=0\)).

Anyway, it has to be pointed out that the Vox2Vox GAN is trained to detected the classes presented in the data and not the combination of them.
\begin{table}[h!]
    \centering
    \caption{Mean dice score and Hausdorff distance 95 percentile for the different brain tumour areas over the training set. The metrics here are obtained training the model with sub-volumes of \(128\times128\times128\) voxels, for different values of \(\alpha=0, 1, 3, 5, 10, 25, 50, 100, 200, 250\).}
    \bgroup
    \def\arraystretch{1.8}
    \begin{tabular}{m{1.2cm}m{1.2cm}m{1.2cm}m{1.2cm}|m{1.2cm}m{1.2cm}m{1cm}}
         & \multicolumn{3}{c}{Dice score [\%]} & \multicolumn{3}{c}{Hausdorff distance 95 [mm]}\\
        \(\alpha\) & WT & TC & ET & WT & TC & ET \\
        \hline
        0   & 58.96 & 23.57 & 37.63 & 77.74 & 104.35 & 73.84 \\
        1   & 86.36 & 72.48 & 62.50 & 13.12 & 17.26 & 40.50 \\
        3   & 92.98 & 90.82 & 79.12 & 4.49 & 4.06 & 31.32\\
        \textbf{5}   & \textbf{93.21} & \textbf{91.70} & \textbf{79.31} & \textbf{3.80} & \textbf{3.18} & \textbf{30.91} \\
        10  & 87.50 & 80.66 & 70.55 & 12.09 & 9.92 & 34.40 \\
        25  & 84.35 & 83.01 & 71.04 & 9.62 & 8.75 & 32.53 \\
        50  & 87.87 & 76.79 & 61.54 & 9.55 & 8.33 & 31.57 \\
        100 & 91.81 & 89.22 & 77.31 & 5.28 & 5.03 & 31.63 \\
        200 & 88.72 & 90.16 & 77.62 & 6.33 & 3.38 & 31.19 \\
        250 & 86.65 & 85.67 & 78.35 & 9.32 & 8.90 & 31.87 \\ 
        \hline
    \end{tabular}
    \label{tab:alphas}
    \egroup
\end{table}
Moreover, it is evident that the ET class is the most problematic one to detect. Indeed, there are just 27 subjects (7.3\%) in the training dataset that do not contain any ET voxels. So, the post-processing should focus on that class in the end, see section 2.6. \\
With the alpha parameter set to 5, 10-fold cross-validation (\(M=10\)) over the training set is applied, using all the image augmentation techniques listed in section 2.2 are applied to avoid overfitting. Every epoch takes approximately 20 minutes to complete using a Keras.utils.Sequence generator. Successively, once the 10 Vox2Vox models are trained, their outputs are combined by the Ensembler CNN explained previously in section 2.6.

As post-processing, a threshold \(th=1000\) voxels (\(V=1\)cm\(^3\)) is set  for the ET class: if the final segmentation has a number of ET voxels fewer than \(th\) ones, the ET voxels are converted into the NCR/NET class. Table \ref{tab:ensemble} reports the metrics calculated on the training, validation and test sets for each class by the CBICA Image Processing Portal\footnote{https://ipp.cbica.upenn.edu, the name of this group for the challenge is IMT\_AE.} with our proposed ensembling and post-processing.
\begin{table}
    \centering
    \caption{Mean, standard deviation and median dice score and Hausdorff distance 95 percentile for the three different brain tumour classes over the training, validation, and test set. The predictions are calculated ensembling 10 models trained after a 10-fold cross-validation and post-processed with a threshold \(th=1000\) voxels for the ET class. The values reported here were calculated by the CBICA Image Processing Portal.}
    \bgroup
    \def\arraystretch{1.8}
    \begin{tabular}{m{2cm}m{2cm}m{1.2cm}m{1.2cm}m{1.2cm}|m{1.2cm}m{1.2cm}m{1cm}}
         & & \multicolumn{3}{c}{Dice score [\%]} & \multicolumn{3}{c}{Hausdorff distance 95 [mm]}\\
         
        dataset &  & WT & TC & ET & WT & TC & ET \\
        \hline
         & Mean & 91.63 & 89.25 & 79.56 & 3.66 & 3.52 & 30.04 \\
        training & StdDev & 6.36 & 11.49 & 24.74 & 3.87 & 4.57 & 96.67 \\
         & Median & 93.39 & 92.50 & 87.16 & 2.44 & 2.23 & 1.73 \\
        \hline
         & Mean & 89.26 & 79.19 & 75.04 & 6.39 & 14.07 & 36.00 \\
        validation & StdDev & 8.28 & 22.30 & 28.68 & 11.44 & 47.56 & 105.28 \\
         & Median & 91.75 & 88.13 & 85.87 & 3.0 & 3.74 & 2.23 \\
        \hline
         & Mean & 87.20 & 81.14 & 78.67 & 6.44 & 24.36 & 18.95\\
        test & StdDev & 14.07 & 25.47 & 20.17 & 11.09 & 79.78 & 74.91\\
         & Median & 91.42 & 90.59 & 83.65 & 3.16 & 3.0 & 2.23\\
        \hline
    \end{tabular}
    \label{tab:ensemble}
    \egroup
\end{table}

The training metrics for the WT and TC class decreased compared to those reported in Table \ref{tab:alphas}, probably due to the image augmentation techniques introduced during the training, but just slightly; on the other hand, the ET ones slightly increased, probably thanks to the ensembling and the post-processing. The metrics obtained for both sets are high, but there are still some bad predictions that compromise the mean values, which also explains the large standard deviations.
For this reason, the median values for each class reported in Table \ref{tab:ensemble} may be more representative of the typical segmentation performance of the model.


\section{Conclusions}
Table \ref{tab:ensemble} establishes that ensembling multiple Vox2Vox models generates high quality segmentation outputs that looks realistic thanks to their discriminators' support, achieving  median values of: 93.40\%, 92.49\%, and 86.48\% dice scores and 2.44mm, 2.23mm, and 1.73mm Hausdorff distance 95 percentile over the training dataset; 91.75\%, 88.13\%, and 83.14\% and 3.0mm, 3.74mm, and 2.44mm over the validation dataset; and 91.42\%, 90.59\%, and 83.65\% and 3.16mm, 3.0mm and 2.23mm over the testing dataset for whole tumour, core tumour and enhancing tumour respectively. 

Moreover, as future work, the Vox2Vox can be improved for the following BraTS challenges in many ways, i.e.: training the model to optimize the BraTS metrics and not those provided in the data, optimizing the PatchGAN architecture, and ensembling Vox2Vox trained with different augmentation techniques, following the suggestions reported in \cite{cirillo2020best}.

In the end, the Vox2Vox model can be used not only for image segmentation but also for further image augmentation. Indeed, Vox2Vox could be combined with a 3D noise to image GAN \cite{eklund,kwon2019generation} which can synthesize realistic segmentation outputs, that are then translated to realistic MR volumes. The combination of these two GANs might result in a really fast batch generation of MR images with their targets. 

\section*{Acknowledgement}
This study was supported by LiU Cancer, VINNOVA Analytic Imaging Diagnostics Arena (AIDA), and the ITEA3 / VINNOVA funded project Intelligence based iMprovement of Personalized treatment And Clinical workflow supporT (IMPACT). Funding was also provided by the Center for Industrial Information Technology (CENIIT) at Linköping University.
\bibliographystyle{splncs04}
\bibliography{mybibliography}

\begin{thebibliography}{10}
\providecommand{\url}[1]{\texttt{#1}}
\providecommand{\urlprefix}{URL }
\providecommand{\doi}[1]{https://doi.org/#1}

\bibitem{bakas3}
Bakas, S., Akbari, H., Sotiras, A., Bilello, M., Rozycki, M., Kirby, J.,
  Freymann, J., Farahani, K., Davatzikos, C.: Segmentation labels and radiomic
  features for the pre-operative scans of the {TCGA-GBM} collection. {T}he
  {C}ancer {I}maging {A}rchive (2017)

\bibitem{bakas4}
Bakas, S., Akbari, H., Sotiras, A., Bilello, M., Rozycki, M., Kirby, J.,
  Freymann, J., Farahani, K., Davatzikos, C.: Segmentation labels and radiomic
  features for the pre-operative scans of the {TCGA-LGG} collection. The Cancer
  Imaging Archive  (2017)

\bibitem{bakas1}
Bakas, S., Akbari, H., Sotiras, A., Bilello, M., Rozycki, M., Kirby, J.S.,
  Freymann, J.B., Farahani, K., Davatzikos, C.: Advancing the cancer genome
  atlas glioma {MRI} collections with expert segmentation labels and radiomic
  features. Scientific data  \textbf{4},  170117 (2017)

\bibitem{bakas2}
Bakas, S., Reyes, M., Jakab, A., Bauer, S., Rempfler, M., Crimi, A., Shinohara,
  R.T., Berger, C., Ha, S.M., Rozycki, M., et~al.: Identifying the best machine
  learning algorithms for brain tumor segmentation, progression assessment, and
  overall survival prediction in the {BRATS} challenge. arXiv preprint
  arXiv:1811.02629  (2018)

\bibitem{blystad}
Blystad, I.: Clinical {A}pplications of {S}ynthetic {MRI} of the {B}rain,
  vol.~1600. Link{\"o}ping University Electronic Press (2017)

\bibitem{cirillo2020best}
Cirillo, M.D., Abramian, D., Eklund, A.: What is the best data augmentation
  approach for brain tumor segmentation using 3d u-net? arXiv preprint
  arXiv:2010.13372  (2020)

\bibitem{dong}
Dong, X., Lei, Y., Wang, T., Thomas, M., Tang, L., Curran, W.J., Liu, T., Yang,
  X.: Automatic multiorgan segmentation in thorax {CT} images using
  {U}-net-{GAN}. Medical physics  \textbf{46}(5),  2157--2168 (2019)

\bibitem{eklund}
Eklund, A.: Feeding the zombies: {S}ynthesizing brain volumes using a 3{D}
  progressive growing {GAN}. arXiv preprint arXiv:1912.05357  (2019)

\bibitem{gatys}
Gatys, L.A., Ecker, A.S., Bethge, M.: Image style transfer using convolutional
  neural networks. In: Proceedings of the IEEE conference on computer vision
  and pattern recognition. pp. 2414--2423 (2016)

\bibitem{goodfellow}
Goodfellow, I., Pouget-Abadie, J., Mirza, M., Xu, B., Warde-Farley, D., Ozair,
  S., Courville, A., Bengio, Y.: Generative adversarial nets. In: Advances in
  neural information processing systems. pp. 2672--2680 (2014)

\bibitem{han}
Han, Z., Wei, B., Mercado, A., Leung, S., Li, S.: Spine-{GAN}: {S}emantic
  segmentation of multiple spinal structures. Medical image analysis
  \textbf{50},  23--35 (2018)

\bibitem{he}
He, K., Zhang, X., Ren, S., Sun, J.: Delving deep into rectifiers: Surpassing
  human-level performance on imagenet classification. In: Proceedings of the
  IEEE international conference on computer vision. pp. 1026--1034 (2015)

\bibitem{he2016deep}
He, K., Zhang, X., Ren, S., Sun, J.: Deep residual learning for image
  recognition. In: Proceedings of the IEEE conference on computer vision and
  pattern recognition. pp. 770--778 (2016)

\bibitem{isensee}
Isensee, F., Kickingereder, P., Wick, W., Bendszus, M., Maier-Hein, K.H.: No
  new-net. In: International MICCAI Brainlesion Workshop. pp. 234--244.
  Springer (2018)

\bibitem{isola}
Isola, P., Zhu, J.Y., Zhou, T., Efros, A.A.: Image--to--image translation with
  conditional adversarial networks. In: Proceedings of the IEEE conference on
  computer vision and pattern recognition. pp. 1125--1134 (2017)

\bibitem{kamnitsas}
Kamnitsas, K., Bai, W., Ferrante, E., McDonagh, S., Sinclair, M., Pawlowski,
  N., Rajchl, M., Lee, M., Kainz, B., Rueckert, D., et~al.: Ensembles of
  multiple models and architectures for robust brain tumour segmentation. In:
  International {MICCAI} {B}rainlesion {W}orkshop. pp. 450--462. Springer
  (2017)

\bibitem{karras}
Karras, T., Aila, T., Laine, S., Lehtinen, J.: Progressive growing of {GANs}
  for improved quality, stability, and variation. ICLR  (2018)

\bibitem{kingma}
Kingma, D.P., Ba, J.: Adam: A method for stochastic optimization. arXiv
  preprint arXiv:1412.6980  (2014)

\bibitem{kong}
Kong, X., Sun, G., Wu, Q., Liu, J., Lin, F.: Hybrid pyramid {U-N}et model for
  brain tumor segmentation. In: International Conference on Intelligent
  Information Processing. pp. 346--355. Springer (2018)

\bibitem{kwon2019generation}
Kwon, G., Han, C., Kim, D.s.: {Generation of 3D brain MRI using auto-encoding
  generative adversarial networks}. In: International Conference on Medical
  Image Computing and Computer-Assisted Intervention. pp. 118--126 (2019)

\bibitem{li}
Li, Y., Shen, L.: c{C-GAN}: {A} robust transfer-learning framework for {HE}p-2
  specimen image segmentation. IEEE Access  \textbf{6},  14048--14058 (2018)

\bibitem{mckinley}
McKinley, R., Meier, R., Wiest, R.: Ensembles of densely-connected {CNN}s with
  label-uncertainty for brain tumor segmentation. In: International MICCAI
  Brainlesion Workshop. pp. 456--465. Springer (2018)

\bibitem{menze}
Menze, B.H., Jakab, A., Bauer, S., Kalpathy-Cramer, J., Farahani, K., Kirby,
  J., Burren, Y., Porz, N., Slotboom, J., Wiest, R., et~al.: The multimodal
  brain tumor image segmentation benchmark ({BRATS}). IEEE transactions on
  medical imaging  \textbf{34}(10),  1993--2024 (2014)

\bibitem{milletari}
Milletari, F., Navab, N., Ahmadi, S.A.: V-net: Fully convolutional neural
  networks for volumetric medical image segmentation. In: 2016 Fourth
  International Conference on 3D Vision (3DV). pp. 565--571. IEEE (2016)

\bibitem{myronenko}
Myronenko, A.: 3{D} {MRI} brain tumor segmentation using autoencoder
  regularization. In: International MICCAI Brainlesion Workshop. pp. 311--320.
  Springer (2018)

\bibitem{nema}
Nema, S., Dudhane, A., Murala, S., Naidu, S.: Rescue{N}et: {A}n unpaired {GAN}
  for brain tumor segmentation. Biomedical Signal Processing and Control
  \textbf{55},  101641 (2020)

\bibitem{polly}
Polly, F., Shil, S., Hossain, M., Ayman, A., Jang, Y.: Detection and
  classification of {HGG and LGG} brain tumor using machine learning. In: 2018
  International Conference on Information Networking (ICOIN). pp. 813--817.
  IEEE (2018)

\bibitem{ronneberger}
Ronneberger, O., Fischer, P., Brox, T.: U-net: Convolutional networks for
  biomedical image segmentation. In: International Conference on Medical image
  computing and computer-assisted intervention. pp. 234--241. Springer (2015)

\bibitem{sato}
Sato, M., Hotta, K., Imanishi, A., Matsuda, M., Terai, K.: {Segmentation of
  Cell Membrane and Nucleus by Improving Pix2pix.} In: BIOSIGNALS. pp. 216--220
  (2018)

\bibitem{sudre}
Sudre, C.H., Li, W., Vercauteren, T., Ourselin, S., Cardoso, M.J.: Generalised
  dice overlap as a deep learning loss function for highly unbalanced
  segmentations. In: Deep learning in medical image analysis and multimodal
  learning for clinical decision support, pp. 240--248. Springer (2017)

\bibitem{topol}
Topol, E.J.: High-performance medicine: the convergence of human and artificial
  intelligence. Nature medicine  \textbf{25}(1),  44--56 (2019)

\bibitem{ulyanov}
Ulyanov, D., Vedaldi, A., Lempitsky, V.: Instance normalization: {T}he missing
  ingredient for fast stylization. arXiv preprint arXiv:1607.08022  (2016)

\bibitem{wesseling}
Wesseling, P., Capper, D.: Who 2016 classification of gliomas. Neuropathology
  and applied neurobiology  \textbf{44}(2),  139--150 (2018)

\bibitem{yi}
Yi, X., Walia, E., Babyn, P.: Generative adversarial network in medical
  imaging: {A} review. Medical image analysis p. 101552 (2019)

\bibitem{zhu}
Zhu, J.Y., Park, T., Isola, P., Efros, A.A.: Unpaired image-to-image
  translation using cycle-consistent adversarial networks. In: Proceedings of
  the IEEE international conference on computer vision. pp. 2223--2232 (2017)

\end{thebibliography}

\end{document}